\documentclass[runningheads]{llncs}

\usepackage[T1]{fontenc}
\usepackage{graphicx}
\usepackage{amsmath}
\usepackage{xcolor}
\usepackage{amssymb}
\usepackage{hyperref}
\usepackage{subcaption}
\usepackage{float}
\usepackage{cite} 

\begin{document}
\setlength{\textfloatsep}{6pt plus 1pt minus 1pt}
\setlength{\floatsep}{6pt plus 1pt minus 1pt}
\setlength{\intextsep}{6pt plus 1pt minus 1pt}
\setlength{\textfloatsep}{10pt plus 1.0pt minus 2.0pt}
\title{Dynamic Class-Aware Active Learning for Unbiased Satellite Image Segmentation}
\titlerunning{DCAU: Active Learning for Imbalanced Satellite Segmentation}

\author{Gadi Hemanth Kumar\inst{1} \and Athira Nambiar\inst{1*} \and Pankaj Bodani\inst{2}}
\authorrunning{G. Hemanth Kumar \and A. Nambiar \and P. Bodani}

\institute{Department of Computational Intelligence, Faculty of Engineering and Technology,\\
SRM Institute of Science and Technology, Kattankulathur 603203, Tamil Nadu, India\\
\email{\{hg7086, athiram\}@srmist.edu.in}
\and
Space Applications Centre, Indian Space Research Organisation, Ahmedabad, India\\
\email{pankajb@sac.isro.gov.in}
}

\maketitle

\begin{abstract}
Semantic segmentation of satellite imagery plays a vital role in land cover mapping and environmental monitoring. However, annotating large-scale, high-resolution satellite datasets is costly and time consuming, especially when covering vast geographic regions. Instead of randomly labeling data or exhaustively annotating entire datasets, Active Learning (AL) offers an efficient alternative by intelligently selecting the most informative samples for annotation with the help of Human-in-the-loop (HITL), thereby reducing labeling costs while maintaining high model performance. AL is particularly beneficial for large-scale or resource-constrained satellite applications, as it enables high segmentation accuracy with significantly fewer labeled samples. Despite these advantages, standard AL strategies typically rely on global uncertainty or diversity measures and lack the adaptability to target underperforming or rare classes as training progresses, leading to bias in the system. To overcome these limitations, we propose a novel adaptive acquisition function, \textbf{Dynamic Class-Aware Uncertainty based Active learning (DCAU-AL)} that prioritizes sample selection based on real-time class-wise performance gaps, thereby overcoming class-imbalance issue.  The proposed DCAU-AL mechanism continuously tracks the performance of the segmentation per class and dynamically adjusts the sampling weights to focus on poorly performing or underrepresented classes throughout the active learning process. Extensive experiments on the OpenEarth land cover dataset show that DCAU-AL significantly outperforms existing AL methods, especially under severe class imbalance, delivering superior per-class IoU and improved annotation efficiency.

\end{abstract}

\vspace{-0.5cm}
\section{Introduction}
Semantic segmentation\cite{shan2023boosting,scala2024semantic,shan2023incremental} of satellite imagery has emerged as a fundamental task in Earth Observation applications such as land cover mapping~\cite{xia2023openearthmap}, disaster response, urban development, and environmental monitoring. Recent advances in deep learning have significantly improved semantic segmentation, enabling models to achieve high pixel-wise accuracy by assigning semantic labels to each pixel and generating fine-grained land cover maps from satellite imagery. However, these models require pixel-level annotations during training, which poses a major challenge. Even with the assistance of auxiliary annotation tools, creating such detailed annotations remains expensive and time consuming, often requires domain experts to meticulously annotate complex geospatial features in large-scale, high-resolution imagery.\\[1ex]
\setlength{\parskip}{0.5ex}
\setlength{\parindent}{0pt}
Active Learning (AL) has emerged as a promising solution to address the annotation burden in semantic segmentation tasks by strategically selecting the most informative samples for labeling with the help of an expert. This human-in-the-loop approach~\cite{mandalika2024segxal} enables models to achieve competitive performance while requiring significantly fewer labeled samples, thereby substantially reducing annotation costs. This methodology proves particularly valuable for large-scale satellite image applications, where comprehensive pixel-level annotation is often prohibitively expensive and time-consuming and also requires domain expert. 

Existing active learning (AL) approaches can be broadly categorized into three categories. \textbf{Uncertainty-based methods} focus on identifying samples with high predictive uncertainty by estimating posterior class probabilities, decision margins, or information entropy~\cite{mittal2025realistic, mandalika2024segxal, yang2015multiclass}. Notable techniques in this category include Monte Carlo dropout~\cite{gal2017deep}, which quantifies model uncertainty to guide efficient sample selection. \textbf{Diversity-based methods}~\cite{mittal2025realistic,agarwal2020contextual,sener2017coreset} prioritize the selection of samples that provide comprehensive coverage across the feature space, thereby ensuring representativeness of the labeled dataset. Finally, \textbf{hybrid methods}~\cite{li2023halia} synergistically combine uncertainty and diversity criteria to simultaneously enhance both the informativeness and representational coverage of selected samples.

\vspace{0.1cm}
 Despite their effectiveness, these strategies encounter substantial difficulties in semantic segmentation tasks with severe class imbalance a prevalent problem in satellite imagery where rare classes such as \textit{bareland} and \textit{water bodies} are significantly outnumbered by dominant categories like vegetation and urban areas, as highlighted in the OpenEarthMap land cover dataset~\cite{xia2023openearthmap}. Traditional uncertainty-based active learning methods tend to exhibit bias toward majority classes, further exacerbating class imbalance across active learning cycles. Conventional class balancing techniques, such as reweighting samples based on the inverse of class frequency~\cite{huang2016learning} or resampling rare class instances~\cite{he2009learning}, are often applied at the data level. However, these methods fail to account for dynamic model behavior and evolving class-wise performance during iterative training. Moreover, they may amplify label noise and introduce sampling bias, particularly in high-resolution satellite imagery with complex scenes. Therefore, a more adaptive approach is needed one that not only acknowledges class imbalance but actively responds to it during learning.

\vspace{0.1cm}
To address the challenges in semantic segmentation under class imbalance, we propose a novel acquisition strategy called \textbf{Dynamic Class-Aware Uncertainty based Active learning (DCAU-AL)}. DCAU-AL adaptively integrates class-wise performance feedback into the active learning loop to improve per-class performance and annotation efficiency. Unlike traditional uncertainty-based strategies that often prioritize dominant classes, DCAU-AL dynamically prioritizes underperforming or underrepresented semantic classes by re-weighting uncertainty scores based on real-time class-wise Intersection-over-Union (IoU) feedback during training. This targeted sampling helps reduce bias and improves segmentation quality across all classes.We conduct extensive experiments on the OpenEarthMap land cover dataset~\cite{xia2023openearthmap}, and evaluate performance using standard metrics such as mean Intersection-over-Union (mIoU) and per-class IoU. Results demonstrate that DCAU-AL significantly outperforms existing active learning strategies in both annotation efficiency and class-wise accuracy, particularly under severe class imbalance conditions. The key contributions of this paper are:
\begin{itemize}
    \item Proposal of a novel \textbf{Dynamic Class-Aware Uncertainty based Active Learning} (DCAU-AL)framework to address the class imbalance problem in semantic segmentation of satellite imagery.
    \item Incorporation of an \textbf{adaptive thresholding mechanism} to balance exploration (sampling diverse new data) and exploitation (focusing on poorly performing classes) during sample acquisition.
    \item Extensive experimental validation on the \textbf{OpenEarthMap} dataset, demonstrating superior annotation efficiency and improved class-wise segmentation performance compared to baseline active learning methods.
\end{itemize}
\vspace{-0.5cm}
\section {Related Work}
\vspace{-0.3cm}
\subsection{Active Learning for Semantic Segmentation}

Several active learning strategies have been proposed to reduce annotation costs in semantic segmentation by targeting different levels of sample selection. Difficulty-awarE Active Learning (DEAL)~\cite{xie2020deal} incorporates semantic difficulty scores to identify informative samples at the image level, while ViewAL~\cite{siddiqui2020viewal} exploits inconsistencies across multi-view predictions for superpixel-level uncertainty estimation. DIAL~\cite{lenczner2022dial} adopts a dual-input scheme combining image and feature space representations for better uncertainty inference and sample efficiency. S4GAN~\cite{mittal2025realistic} combines self-supervised pretraining and semi-supervised consistency regularization with active querying, delivering strong performance under realistic annotation budgets by leveraging both labeled and unlabeled data. More recently, Plug-and-Play Active Learning (PPAL)~\cite{yang2024ppal} introduced a two-stage, architecture-agnostic strategy for object detection, integrating difficulty-calibrated uncertainty sampling with category-conditioned similarity matching, and demonstrated significant performance gains under limited labeling budgets.

\subsection{Class Imbalance in Active Learning}

Class imbalance is an ill-posed problem ~\cite{japkowicz2000class}, particularly in semantic segmentation, where rare classes occupy fewer pixels, thus reducing their influence during training and uncertainty estimation. Common approaches to address this issue include reweighting~\cite{ren2018learning} and resampling~\cite{he2009learning}. Reweighting modifies the loss function by assigning higher weights to minority classes, often based on the frequency of the inverse class~\cite{aggarwal2020active}, with further refinements such as the effective number of samples~\cite{cui2018large}. Resampling adjusts the training distribution through over- or under-sampling~\cite{he2009learning}, while more recent work has investigated robust reweighting strategies under label noise~\cite{ren2018learning}.

\vspace{0.1cm}
In the context of Active Learning (AL), most existing methods apply \textit{static class weights}, which fail to adapt to evolving model performance. For example, Aggarwal et al.~\cite{aggarwal2020active} used historical class frequencies to rebalance sample selection in classification tasks, while Barata et al.~\cite{barata2021coldstart} proposed a staged AL pipeline to mitigate cold-start issues. However, these methods are designed for classification and do not translate well to semantic segmentation. To address this, Shan et al.~\cite{shan2024edge} introduced an AL framework for aerial image segmentation using edge-guided label units and static class balancing. While promising, it does not adapt dynamically to changing class distributions and depends on pseudo-labels, which can introduce noise and limit effectiveness.
\footnote{OpenEarth dataset class distribution: Bareland 1.5\%, Rangeland 22.9\%, Developed space 16.1\%, Road 6.7\%, Tree 20.2\%, Water 3.3\%, Agriculture land 13.7\%, Building 15.6\%.}

\vspace{0.1cm}
In contrast to existing approaches, our proposed method addresses these limitations by dynamically adjusting sampling weights in real-time through a novel \textbf{Dynamic Class-Aware Uncertainty (DCAU)} technique, thereby enhancing sampling efficiency and improving performance on underrepresented classes. To the best of our knowledge, this work is the first to (i) integrate \textbf{class-wise performance feedback} into the active learning loop for semantic segmentation, and (ii) introduce an \textbf{adaptive, uncertainty-driven selection mechanism} that balances both rare and frequent classes during sample acquisition.

\vspace{-0.3cm}
\section{Methodology: Dynamic Class Aware Uncertainty based Active Learning (DCAU)}
\vspace{-0.5cm}
Active Learning (AL) \textbf{is} an effective strategy to reduce annotation costs by iteratively selecting the most informative samples for labeling, with a human in the loop~\cite{mandalika2024segxal} who annotates only the most valuable samples. AL follows an iterative process~\cite{ren2021survey}: an initial labeled dataset \(D_L\) is used to train the model, while the remaining unlabeled data forms a pool \(D_U\). Based on the model’s predictions, a subset \( \mathcal{S} \subset \mathcal{D}_U \) containing the most informative or uncertain samples is selected for annotation in each round. The newly labeled samples are then added to \(D_L\), and the model is retrained, repeating the cycle until a desired performance level is reached. This human-in-the-loop process significantly reduces labeling costs, making AL especially effective for large-scale satellite image applications where exhaustive pixel-level annotation is impractical.  

However, during the sample selection process, the model often exhibits bias towards dominant classes. In semantic segmentation tasks, this becomes particularly problematic due to severe class imbalance, where rare classes are significantly underrepresented. In the case of satellite image semantic segmentation, such as with the OpenEarth Land Cover dataset~\cite{xia2023openearthmap}, models tend to prioritize majority classes such as agricultural land, developed space and rangeland while minority classes such as bareland, roads, and water are often underrepresented. As a result, traditional uncertainty-based sampling methods tend to select regions dominated by majority classes, leading to suboptimal segmentation performance for rare but crucial classes.
\vspace{.2cm}

\begin{figure}[t!]
    \centering
    \includegraphics[width=1\textwidth]{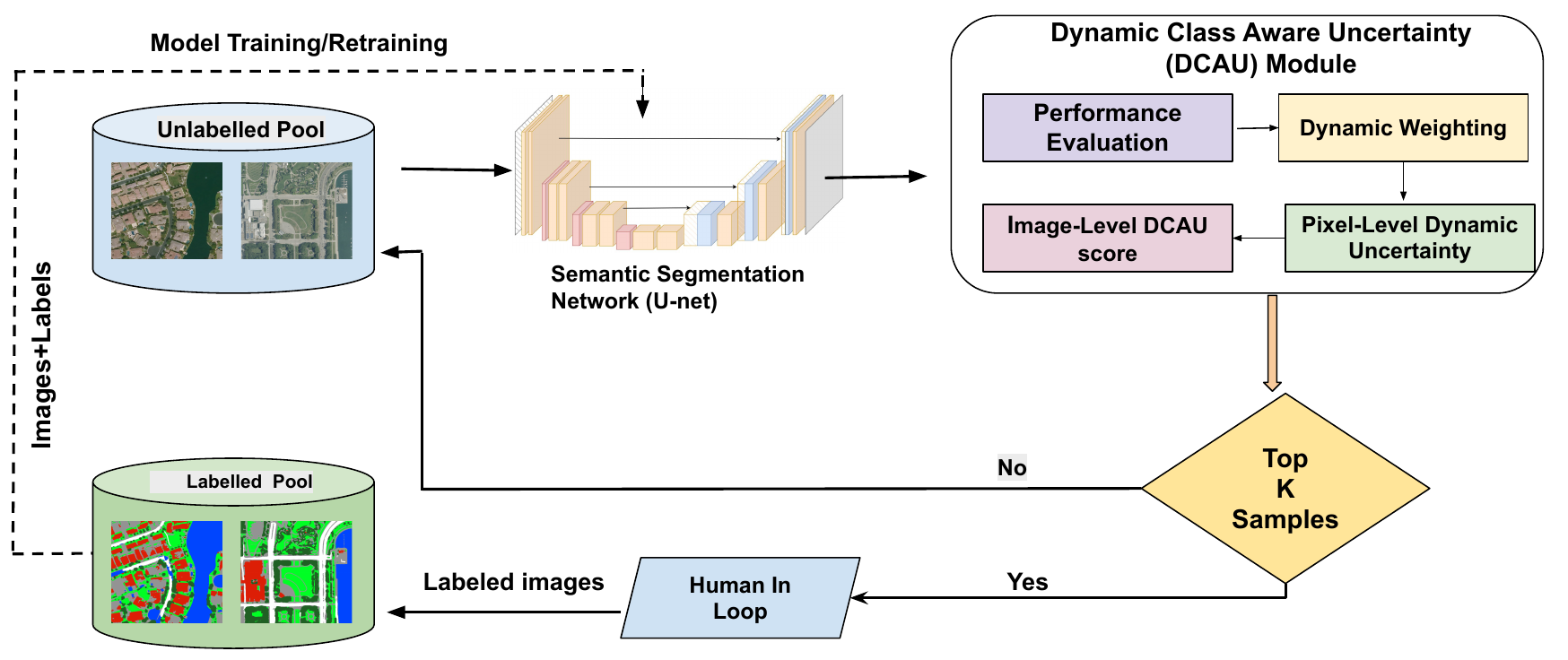}
    \caption{Proposed Dynamic Class Aware Uncertainty based Active Learning (DCAU-AL)  framework for semantic segmentation with a human-in-the-loop.}
    \label{fig:dcau-framework}
\end{figure}

To address the class imbalance limitations of existing active learning approaches, we propose a novel \textbf{Dynamic Class-Aware Uncertainty based AL (DCAU-AL) framework }, as illustrated in Fig.~\ref{fig:dcau-framework}. This novel acquisition strategy dynamically adjusted sampling weights based on per-class performance during each Active Learning cycle, ensuring that underperforming and underrepresented classes receive adequate attention throughout the learning process.
\vspace{.2cm}

The proposed DCAU-based Active Learning (DCAU-AL) framework, illustrated in Fig.~\ref{fig:dcau-framework}, operates through a systematic workflow that begins with an \textbf{unlabeled pool} of satellite images. These images are fed into a semantic segmentation network (UNet in our implementation ~\cite{ronneberger2015unet}) trained on the current \textbf{labeled pool}. The network generates pixel-level class probability maps for each image, which are then analyzed by the \textbf{Dynamic Class-Aware Uncertainty (DCAU) module} to identify the 
most informative samples. The top-ranked samples are forwarded to the human-in-the-loop annotator, who provides or corrects the labels, thereby expanding the labeled pool. The segmentation model is retrained on this updated dataset, and the cycle is repeated until the annotation budget is exhausted. This iterative workflow ensures that each AL cycle focuses on the most uncertain and underrepresented regions, thereby achieving balanced improvement across classes. The detailed formulation of the DCAU module is presented in the following section.

\subsection*{3.1 Dynamic Class-Aware Uncertainty}
Let $\mathbf{x}_i \in \mathbb{R}^{H \times W \times C}$ denote an image, and let
$\mathbf{y}_i \in \{1, 2, \ldots, K\}^{H \times W}$ be its corresponding
ground-truth segmentation mask, where $K$ is the number of semantic classes.  
The dataset is represented as:
\[
\mathcal{D} = \left\{ \left(\mathbf{x}_i, \mathbf{y}_i \right) \right\}_{i=1}^N.
\]
Given a total annotation budget $B$, the objective is to iteratively select a subset 
$\mathcal{S} \subset \mathcal{D}$ such that $|\mathcal{S}| \leq B$, and to maximize 
segmentation performance, particularly for minority classes, through an active 
learning loop guided by the DCAU module.

At each Active Learning (AL) iteration $t$, the DCAU module first performs 
\textbf{class-wise performance evaluation} on the validation set using the 
Intersection over Union (IoU) metric, defined as:

\begin{equation}
\text{IoU}_c(t) = \frac{\mathrm{TP}_c}{\mathrm{TP}_c + \mathrm{FP}_c + \mathrm{FN}_c},
\label{eq:iou}
\end{equation}

where $\mathrm{TP}_c$, $\mathrm{FP}_c$, and $\mathrm{FN}_c$ denote the true positives, 
false positives, and false negatives for class $c$ at iteration $t$. 
A lower $\text{IoU}_c(t)$ indicates that the model is underperforming on class $c$.

To quantify this behavior, the \textbf{performance gap} for class $c$ at iteration $t$ 
is defined as:

\begin{equation}
\text{Gap}_c(t) = 1 - \text{IoU}_c(t).
\label{eq:gap}
\end{equation}

A larger $\text{Gap}_c(t)$ indicates weaker model performance on that class.

Next, the module assigns \textbf{dynamic weights} to each class based on the computed 
gaps. The weight for class $c$ at iteration $t$ is given by:

\begin{equation}
w_c(t) = \frac{(\text{Gap}_c(t))^\alpha}{\sum_{j=1}^{K} (\text{Gap}_j(t))^\alpha},
\label{eq:weight}
\end{equation}
The weighting formulation in Eq.~(\ref{eq:weight}) is designed to
nonlinearly scale the influence of underperforming classes based on their
performance gaps. A simple linear weighting
(\( w_c \propto \text{Gap}_c \)) provided limited differentiation between
majority and minority classes, whereas an exponential formulation led to
training instability by excessively biasing the model toward the lowest-performing
classes. The adopted power-law formulation introduces a tunable control
parameter \(\alpha\) that allows smooth adjustment of emphasis across classes.
Smaller \(\alpha\) values (\(<1\)) encourage balanced attention, while larger
values (\(>1\)) increase focus on classes with lower IoU scores. Empirical
observations indicated that setting \(\alpha = 0.5\) offers a stable trade-off
between robustness and targeted learning, resulting in consistent improvements
in class-wise segmentation performance.

The framework then computes \textbf{pixel-level uncertainty} using the entropy of 
predicted class probabilities $\mathbf{p}_i$ for each pixel $i$:

\begin{equation}
H(\mathbf{p}_i) = -\sum_{k=1}^{K} p_{i,k} \log(p_{i,k}),
\label{eq:entropy}
\end{equation}

where $p_{i,k}$ is the probability that pixel $i$ belongs to class $k$. 
To emphasize uncertainty for underrepresented or poorly performing classes, 
the dynamic weights $w_c(t)$ from Eq.~\eqref{eq:weight} are applied, yielding the 
\textbf{dynamic pixel uncertainty}:

\begin{equation}
H_{\text{dyn}}(\mathbf{p}_i) = \sum_{k=1}^{K} \big(p_{i,k} \cdot w_k(t) \cdot H(\mathbf{p}_i)\big).
\label{eq:dyn_entropy}
\end{equation}

The \textbf{image-level DCAU score} for an image $\mathbf{x}$ is then obtained by 
averaging the dynamic pixel uncertainties across all $N$ pixels:

\begin{equation}
\text{DCAU}(\mathbf{x}) = \frac{1}{N} \sum_{i=1}^{N} H_{\text{dyn}}(\mathbf{p}_i).
\label{eq:dcau}
\end{equation}

To avoid excessive selection of noisy or outlier samples, 
an \textbf{adaptive threshold}, denoted as $\theta(t)$, is applied:
\begin{equation}
\theta(t) = \mu(t) + \gamma \cdot \sigma(t),
\label{eq:threshold}
\end{equation}

where $\mu(t)$ and $\sigma(t)$ denote the mean and standard deviation of DCAU scores across the unlabeled pool at iteration $t$, while $\gamma$ is a scaling factor that controls the strictness of sample selection.

The rationale for introducing $\gamma$ is to avoid overselecting noisy or outlier samples that may have high uncertainty but contain limited useful information. When $\gamma$ is set to a small value, the threshold $\theta(t)$ becomes less selective, allowing more samples (including potentially noisy ones) to be chosen. In contrast, a higher $\gamma$ value raises the selection threshold, focusing only on the most confidently uncertain samples and ensuring that the selected samples are both reliable and informative. In essence, $\gamma$ acts as a stability regulator that adaptively adjusts the balance between exploration and exploitation between iterations.

Empirical observations indicate that setting $\gamma = 0.5$ provides an optimal trade-off, promoting exploration during early active learning cycles and gradually tightening the selection criterion as the model becomes more confident. Furthermore, $\gamma$ can be treated as a \emph{learnable parameter} that dynamically adapts based on model feedback or validation performance, enabling a more flexible and data-driven regulation of the sample selection process. This mechanism maintains consistent performance gains while preventing drift caused by unstable or redundant sample acquisitions.

From these candidates, the \textbf{Top-$K$ samples} are chosen and sent to the 
\textbf{human-in-the-loop} annotator, who labels them and moves them into the labeled pool. 
The segmentation model is then re-trained on this updated dataset, and the process repeats
until the annotation budget $B$ is exhausted. This iterative loop ensures that each AL cycle
focuses on both uncertain and underrepresented classes, leading to balanced performance
improvement across all classes.

\vspace{0.2cm}
\section{Experimental Setup}

\textbf{Dataset}: \textbf{OpenEarth Land Cover}~\cite{xia2023openearthmap} dataset, which provides high-resolution satellite imagery annotated for land use and land cover classification. The dataset comprises a total of 4300 samples, each with a spatial resolution of 1024×1024 pixels. For computational efficiency, all images are resized to 512×512. The dataset contains 9 semantic classes: \textit{Background}, \textit{Bareland}, \textit{Road}, \textit{Building}, \textit{Water}, \textit{Developed Space}, \textit{Rangeland}, \textit{Agricultural Land}, and \textit{Tree}. The data is partitioned into a training set (initially 500 labeled samples), validation set (250 samples), and test set (250 samples). The remaining 3300 samples form the unlabeled pool used during active learning cycles.

\vspace{0.15cm}
\textbf{Implementation Details}:We implement our framework using PyTorch and train a U-Net segmentation model~\cite{ronneberger2015unet} with an encoder-decoder architecture and a 9-class softmax output. Input images are resized to 512×512 and normalized using ImageNet statistics. The model is optimized using the Adam optimizer (learning rate: 0.0001, batch size: 8) for 100 epochs per active learning cycle. In each of the 20 iterations, 50 high uncertainty samples, weighted by class performance gaps, are selected using the proposed DCAU strategy.The implementation was conducted on a system equipped with an NVIDIA GeForce RTX 4070 Ti GPU (32 GB VRAM), and the full active learning pipeline takes approximately 30 hours to complete for 20 iterations, depending on the model configuration.

\section{Experimental Results}
In this section, we present a comprehensive evaluation of the proposed dynamic class-aware uncertainty-based active learning (DCAU-AL)  framework. The experiments are designed to assess both the quantitative and qualitative performance of our approach compared to standard baselines under varying annotation budgets.  In addition, we compare our method against fully supervised and naive active learning settings, and conduct ablation studies to evaluate the impact of key hyperparameters and sampling strategies.

\subsubsection*{5.1.1 Quantitative Comparison of Fully Supervised, Naive AL, and Proposed DCAU-AL Models}

We compare the performance of three approaches on the \textbf{OpenEarth Land Cover} dataset~\cite{xia2023openearthmap}: a fully monitored model trained with data labeled $100\%$, a baseline of naive active learning (Naive AL) and the proposed DCAU-AL framework. All experiments start with an initial labeled set of 500 images out of 3800, while the remaining samples 3300 are the unlabeled pool. At each AL cycle, 50 new samples are annotated and added to the training set. Performance is evaluated on a held-out test set using mean Intersection-over-Union (mIoU) and per-class IoU.
\vspace{0.2cm}

Table~\ref{tab:iuoscore} summarizes the performance analysis of all the experiments. The Fully Supervised model achieves an mIoU of $0.670$, representing the upper bound when all labels are available. The Naive AL approach attains $0.642$, but exhibits a clear bias toward majority classes such as \textit{Tree}, and \textit{Water}, while underperforming in minority and structurally complex categories such as \textit{Bareland} and \textit{Road}. In contrast, DCAU-AL achieves $0.664$ using only 40\% of the labeled data, closely matching the Fully Supervised performance and surpassing Naive AL across nearly all classes. Notably, DCAU-AL improves the IoU of \textit{Bareland} and \textit{Road} to $0.248$ and $0.582$, compared to Naive AL’s $0.201$ and $0.547$, respectively.

\begin{table}[t]
    \centering
    \renewcommand{\arraystretch}{1.7}
    \resizebox{\textwidth}{!}{%
    \begin{tabular}{|c|c|c|c|c|c|c|c|c|c|c|}
        \hline
        Method & mIoU & Background & Bareland & Rangeland & Developed Space & Road & Tree & Water & Agriculture Land & Buildings \\ \hline
        Fully Supervised & 0.670 & 0.703 & 0.217 & 0.694 & 0.638 & 0.603 & 0.786 & 0.858 & 0.704 & 0.749 \\ \hline
        Naive AL         & 0.642 & 0.719 & 0.201 & 0.665 & 0.589 & 0.547 & 0.770 & 0.829 & 0.651 & 0.687 \\ \hline
        DCAU-Based AL    & 0.664 & 0.833 & 0.248 & 0.661 & 0.604 & 0.582 & 0.758 & 0.834 & 0.664 & 0.714 \\ \hline
    \end{tabular}%
    }
    \vspace{0.2cm}
    \caption{Comparison of mIoU and class-wise IoU scores for Fully Supervised, Naive AL, and DCAU-Based AL models. DCAU achieves near-Fully Supervised performance using only 40\% labeled data and delivers significant gains in minority classes.}
    \label{tab:iuoscore}
\end{table}

\subsubsection*{5.1.2 Performance Progression of the Proposed DCAU-AL Framework}

To further analyze the effectiveness of DCAU-AL, the mean Intersection over Union (mIoU) and class-wise IoU scores across 20 Active Learning (AL) cycles are reported in Table~\ref{tab:DCAUscore}. The experiment starts with 500 labeled samples and adds 50 new labeled samples per cycle. The model demonstrates a steady performance improvement, with mIoU increasing from $0.584$ at Cycle~1 to $0.664$ at Cycle~20. The largest relative gains occur in minority classes: \textit{Bareland} improves from $0.151$ to $0.248$, and \textit{Road} from $0.489$ to $0.579$, highlighting DCAU’s effectiveness in mitigating class imbalance over time. Majority classes such as \textit{Background} and \textit{Water} also show incremental improvements, ensuring balanced performance across categories (see Table~\ref{tab:DCAUscore}).

\begin{table}[H]
    \centering
    \renewcommand{\arraystretch}{1.4}
    \resizebox{\textwidth}{!}{%
    \begin{tabular}{|c|c|c|c|c|c|c|c|c|c|c|}
        \hline
        Cycle & mIoU & Background & Bareland & Rangeland & Developed Space & Road & Tree & Water & Agriculture Land & Building \\ \hline
        AL-1  & 0.584 & 0.703 & 0.151 & 0.636 & 0.553 & 0.489 & 0.746 & 0.803 & 0.617 & 0.559 \\ \hline
        AL-5  & 0.604 & 0.803 & 0.168 & 0.621 & 0.549 & 0.492 & 0.722 & 0.805 & 0.570 & 0.673 \\ \hline
        AL-10 & 0.631 & 0.828 & 0.171 & 0.643 & 0.563 & 0.542 & 0.741 & 0.824 & 0.605 & 0.676 \\ \hline
        AL-15 & 0.652 & 0.831 & 0.228 & 0.667 & 0.601 & 0.563 & 0.732 & 0.820 & 0.669 & 0.695 \\ \hline
        AL-20 & 0.664 & 0.833 & 0.248 & 0.661 & 0.604 & 0.579 & 0.758 & 0.834 & 0.664 & 0.714 \\ \hline
    \end{tabular}%
    }
    \vspace{0.2cm}
    \caption{DCAU-AL model mIoU and class-wise IoU scores across AL cycles on the OpenEarth dataset starting with 500 labeled samples.}
    \label{tab:DCAUscore}
\end{table}

\begin{figure}
    \centering
    \includegraphics[width=0.65\linewidth]{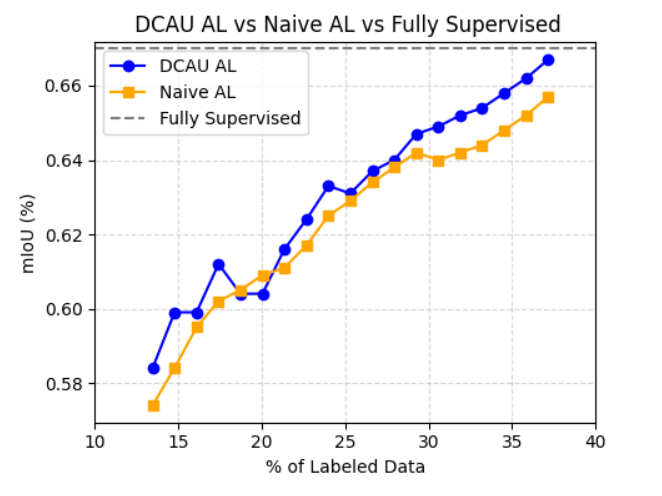}
    \caption{Progression of mIoU scores across acquisition cycles for Fully Supervised, Naive AL, and DCAU-Based AL models.(Better viewed in colour)}
    \label{fig:mio_3}
\end{figure}

This improvement can be attributed to DCAU's dynamic reweighting of uncertainty scores based on performance gaps for each class. This strategy encourages the selection of samples rich in underrepresented or poorly performing classes. By adaptively steering the annotation budget toward classes that require more attention, DCAU-AL maintains overall accuracy while systematically reducing class-specific errors (Table~\ref{tab:DCAUscore}).

Fig~\ref{fig:mio_3} illustrates the progression of mIoU scores for all three compared methods. The Fully Supervised model (dashed line) serves as the upper performance bound, showing a steady improvement as more labeled data becomes available.
DCAU-AL method exhibits consistent convergence toward the upper bound while requiring substantially fewer manual annotations. Notably, DCAU-AL demonstrates superior performance relative to the Naive AL baseline across all evaluation cycles, with the most pronounced performance differential occurring during initial iterations when annotated samples remain limited.
The quantitative results demonstrate that by Cycle 20, DCAU-AL achieves a mean IoU of 0.667 utilizing merely 40\% of the available labeled dataset. This performance level effectively matches that of the Fully Supervised approach while achieving substantial reductions in annotation overhead, as illustrated in Fig~\ref{fig:mio_3}. These findings underscore the efficacy of the proposed uncertainty-aware selection strategy in maximizing model performance under constrained annotation budgets.
\vspace{0.3cm}

\vspace{-0.5cm}
\begin{figure}[htbp]
    \centering
    \begin{subfigure}[b]{0.45\textwidth}
        \includegraphics[width=\textwidth, height=2.8cm]{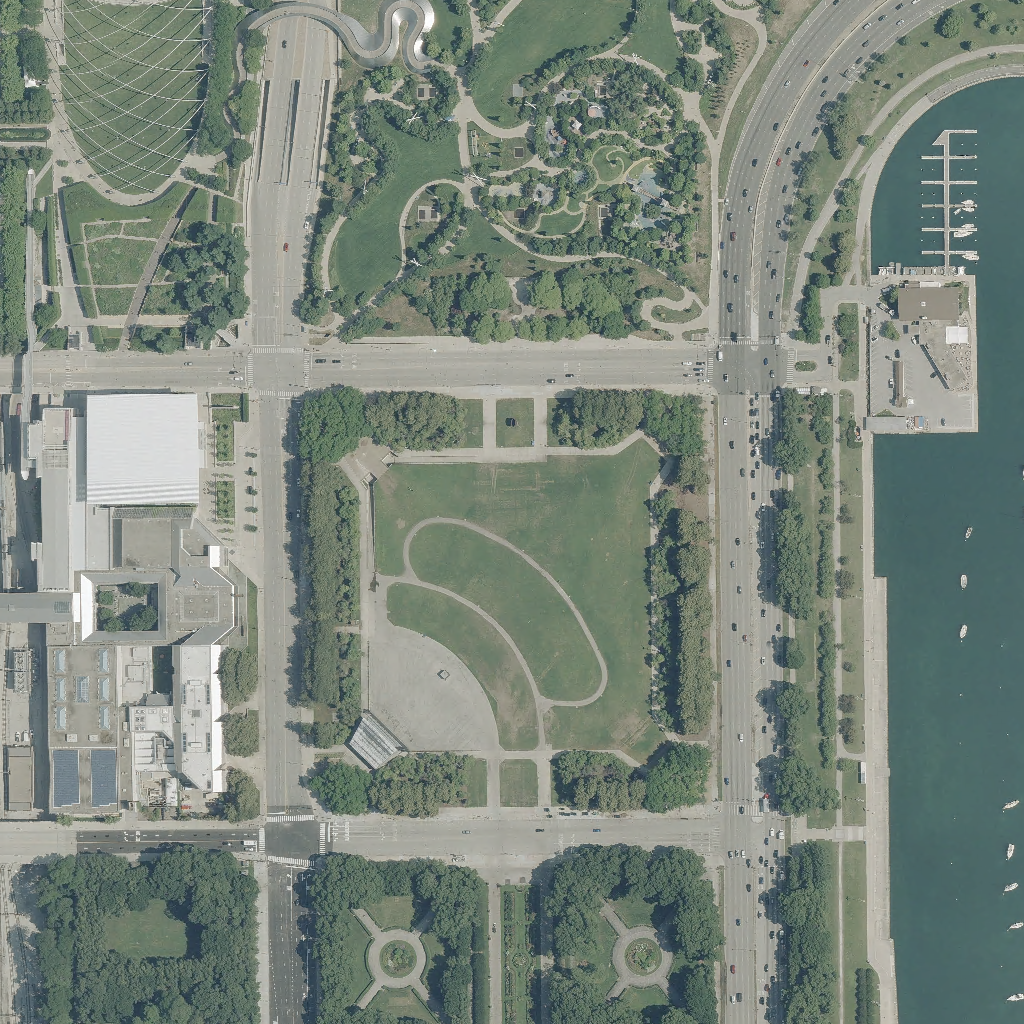}
        \caption*{(a) Raw Image}
    \end{subfigure}
    \hfill
    \begin{subfigure}[b]{0.45\textwidth}
        \includegraphics[width=\textwidth, height=2.8cm]{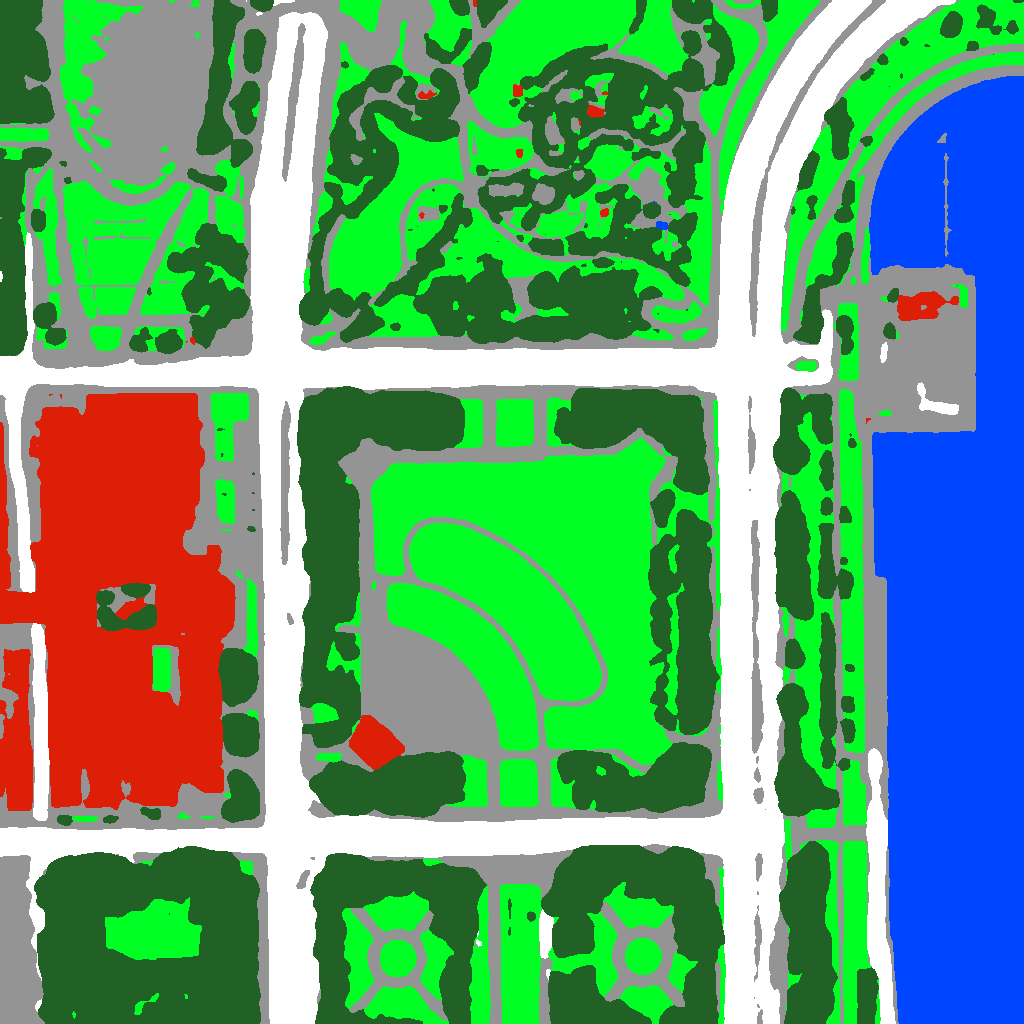}
        \caption*{(b) Ground Truth}
    \end{subfigure}

    \vspace{3mm}
    \begin{subfigure}[b]{0.45\textwidth}
        \includegraphics[width=\textwidth, height=2.8cm]{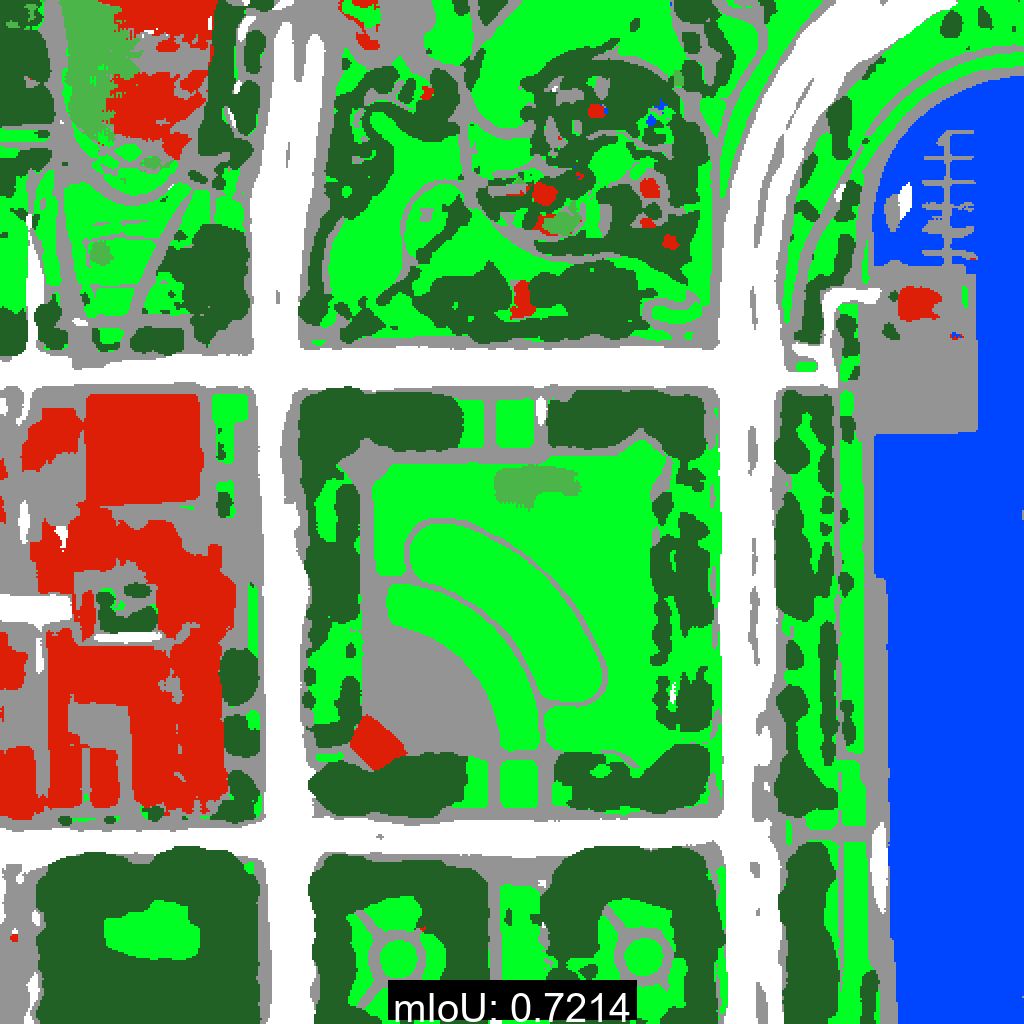}
        \caption*{(c) Fully Supervised (mIoU=0.67)}
    \end{subfigure}
    \hfill
    \begin{subfigure}[b]{0.45\textwidth}
        \includegraphics[width=\textwidth, height=2.8cm]{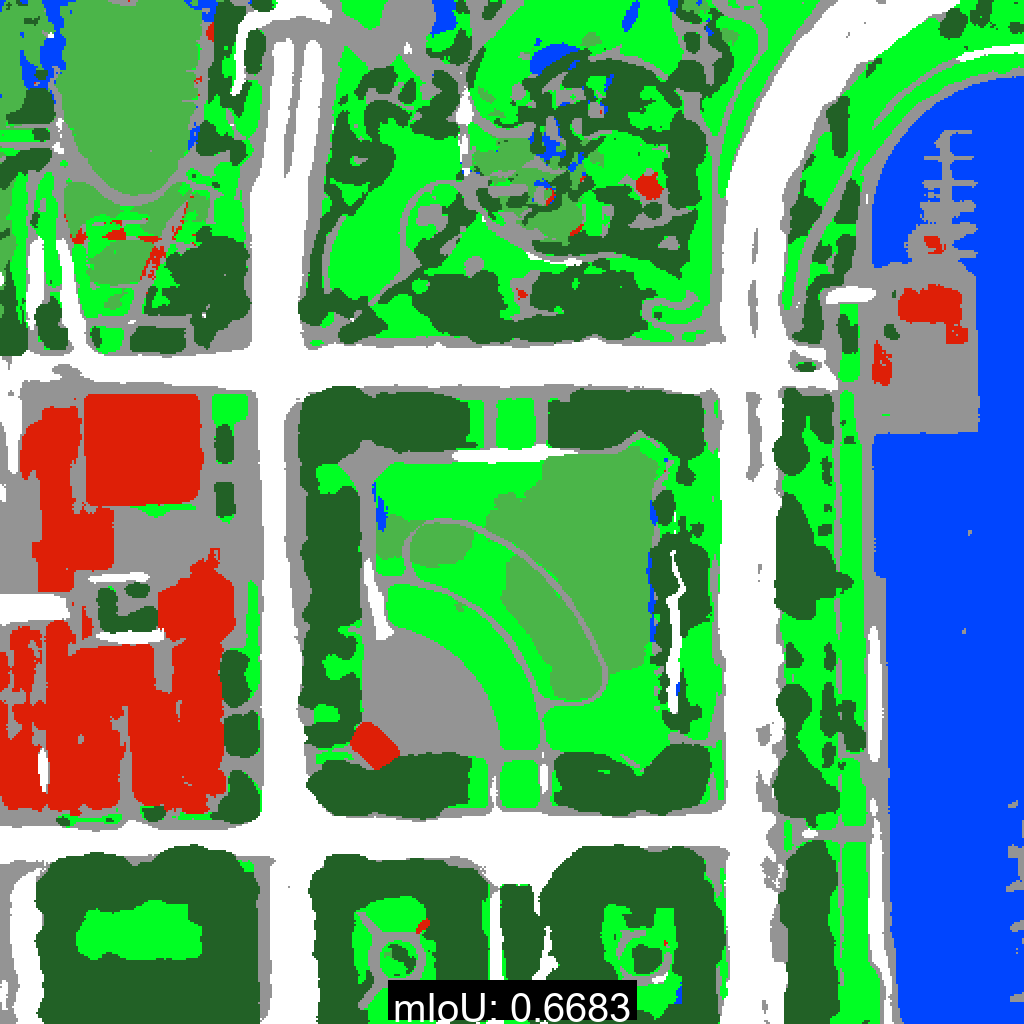}
        \caption*{(d) Naive-AL Cycle  @ 20 (mIoU=0.64)}
    \end{subfigure}

    \vspace{3mm}
    \begin{subfigure}[b]{0.45\textwidth}
        \includegraphics[width=\textwidth, height=2.8cm]{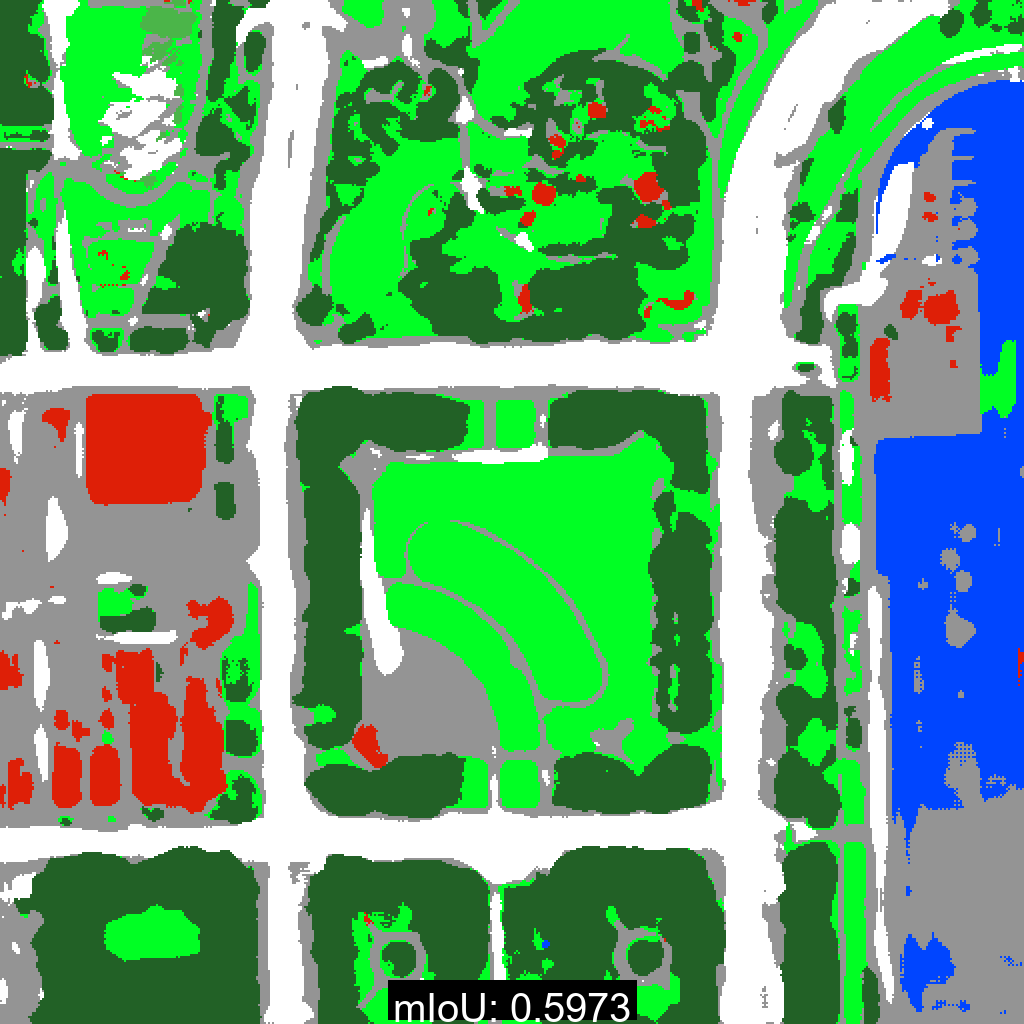}
        \caption*{(e) DCAU-AL Cycle @1 (mIoU=0.58)}
    \end{subfigure}
    \hfill
    \begin{subfigure}[b]{0.45\textwidth}
        \includegraphics[width=\textwidth, height=2.8cm]{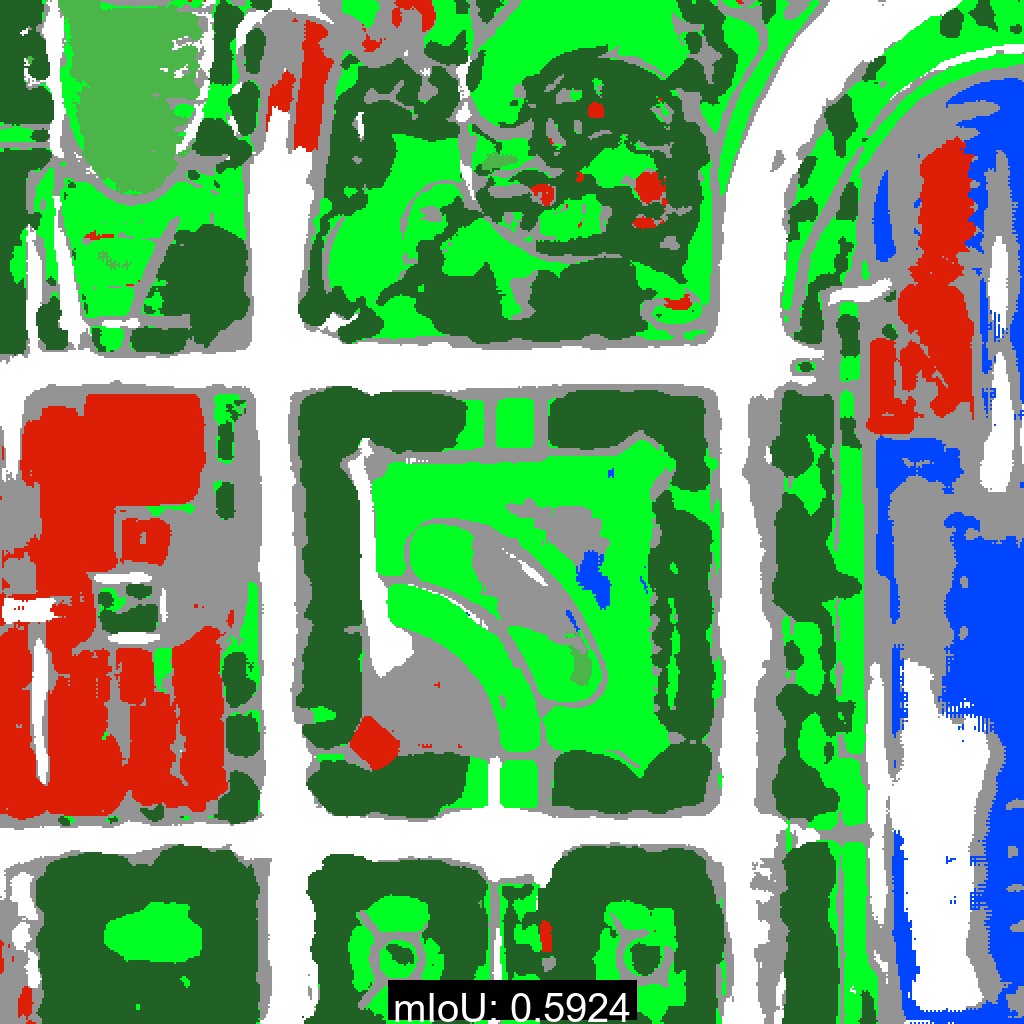}
        \caption*{(f) DCAU-AL Cycle @5 (mIoU=0.59)}
    \end{subfigure}

    \vspace{3mm}
    \begin{subfigure}[b]{0.45\textwidth}
        \includegraphics[width=\textwidth, height=2.8cm]{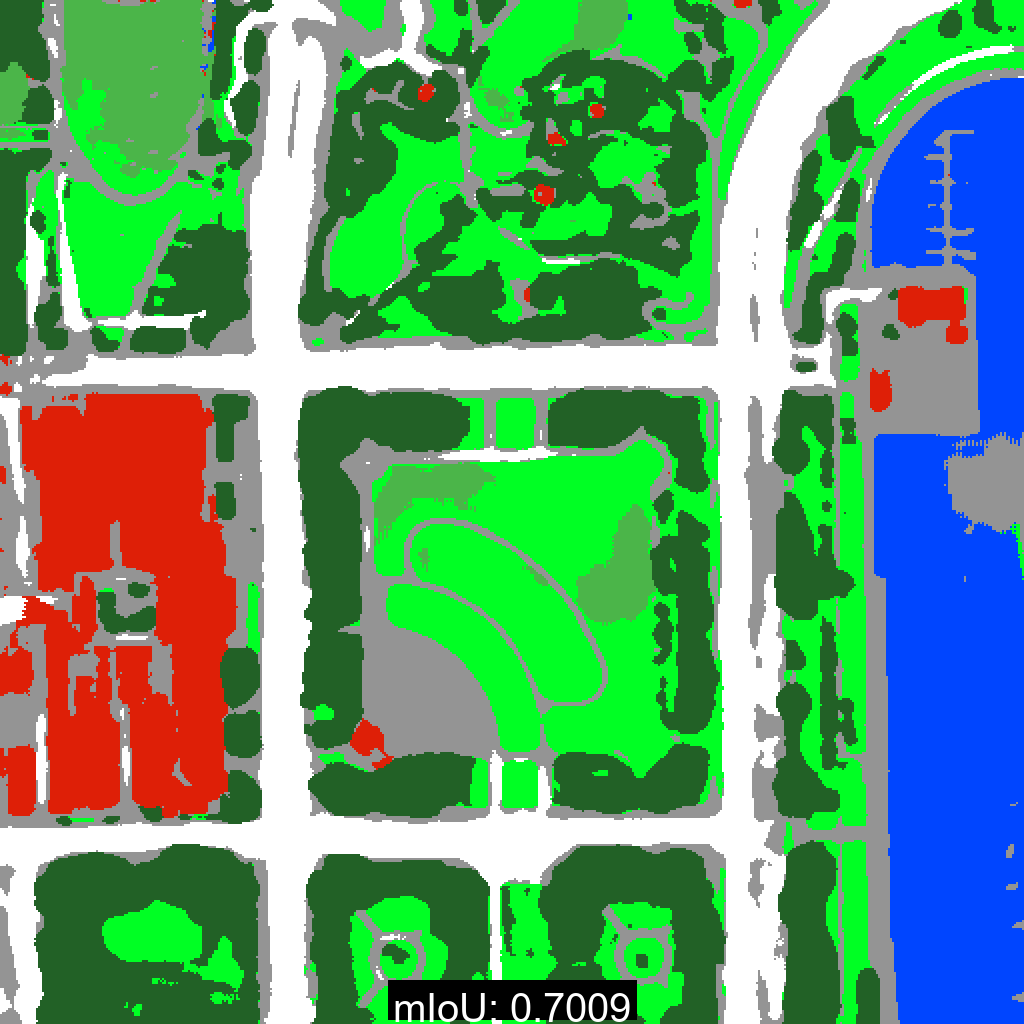}
        \caption*{(g) DCAU-AL Cycle @10 (mIoU=0.63)}
    \end{subfigure}
    \hfill
    \begin{subfigure}[b]{0.45\textwidth}
        \includegraphics[width=\textwidth, height=2.8cm]{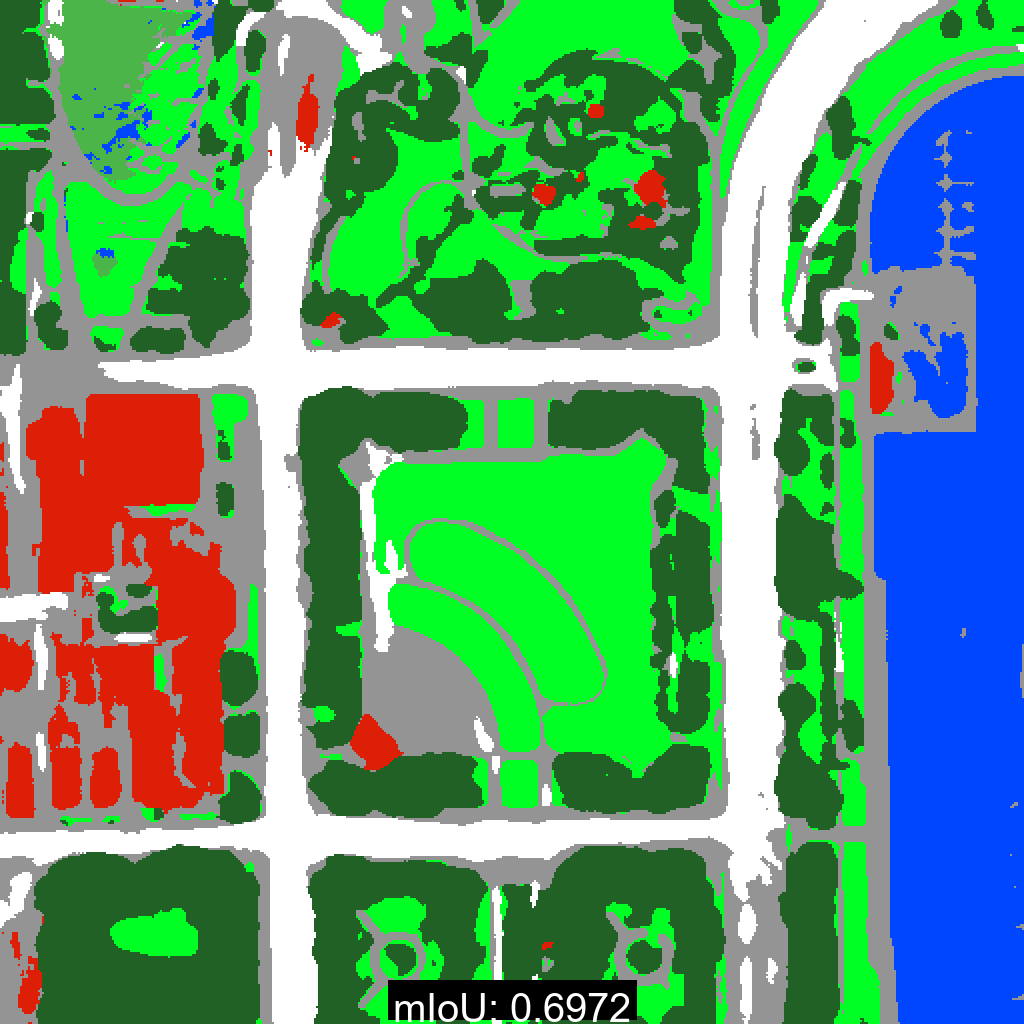}
        \caption*{(h) DCAU-AL Cycle @20 (mIoU=0.66)}
    \end{subfigure}
    \caption{Segmentation results for Fully Supervised, Naive-AL, and DCAU-AL models across AL cycles, with corresponding mIoU values in parentheses.(Better Viewed in Colour)}
    \label{fig:cycle_comparison}
\end{figure}

\subsubsection*{5.2 Qualitative Results}

To demonstrate the efficacy of the proposed DCAU module, qualitative visualizations are also presented  in Fig~\ref{fig:cycle_comparison}. Fig~\ref{fig:cycle_comparison}(a) and Fig~\ref{fig:cycle_comparison}(b) show the raw satellite image and its corresponding ground truth. Fig~\ref{fig:cycle_comparison}(c) and Fig~\ref{fig:cycle_comparison}(d) display the segmentation masks of the Fully Supervised model ($\mathrm{mIoU} = 0.67$) and the Naive-AL model at Cycle~20 ($\mathrm{mIoU} = 0.64$), respectively. The Fully Supervised output captures most fine-grained and minority regions with high accuracy, whereas Naive-AL at Cycle~20 exhibits noticeable class-imbalance errors, particularly in minority classes such as narrow roads and small building areas. 

Fig~\ref{fig:cycle_comparison}(e)--Fig\ref{fig:cycle_comparison}(h) illustrate the DCAU-AL predictions across Active Learning Cycles~1,~5,~10, and~20. DCAU-AL shows consistent improvements across cycles, with minority-class segmentation gradually becoming more precise and less fragmented. By Cycle~20, DCAU-AL achieves $\mathrm{mIoU} = 0.66$, surpassing Naive-AL under the same labeling budget, and captures finer structural details, especially in small buildings, road boundaries, and vegetation edges. This demonstrates that DCAU-AL not only improves efficiency in low-label settings but also mitigates class-imbalance issues over successive cycles, highlighting the robustness of our acquisition strategy in semantic segmentation.

\subsection*{5.3 Ablation Study}

To comprehensively evaluate the robustness and parameter sensitivity of the proposed \textbf{Dynamic Class-Aware Uncertainty (DCAU)} based AL acquisition strategy, we conducted an extensive ablation study by systematically varying key training parameters. Specifically, we analyzed the influence of annotation budget per active learning cycle by increasing the number of labeled samples selected from 50 (baseline configuration) to 100 per iteration, allowing us to assess how sampling granularity affected convergence behavior and class-wise learning efficiency. Furthermore, we investigated the method's sensitivity to optimization settings by evaluating performance under different learning rates: $1 \times 10^{-4}$, and $1\times 10^{-3}$. These systematic variations enabled us to examine the interplay between gradient-based optimization and uncertainty-driven sample acquisition mechanisms.

As shown in Table~\ref{tab:dcau_ablation}, the ablation results reveal important insights about the method's parameter sensitivity. The DCAU-Baseline configuration achieves optimal performance with a learning rate of  $1 \times 10^{-4}$
and 50 samples per AL cycle, yielding an mIoU of 0.664. Interestingly, increasing the learning rate to $1 \times 10^{-3}$ with the same sampling budget results in a slight performance decrease to 0.654. When extending the annotation budget to 100 samples per cycle (DCAU-Extended), the method shows reduced performance across both learning rates, achieving 0.648 with $1 \times 10^{-3}$ and 0.632 with $1 \times 10^{-3}$.  These results suggest that the smaller, more frequent sampling approach (50 samples per cycle) with a conservative learning rate ($1 \times10^{-3}$provides the most effective balance between exploration and exploitation in the active learning process, validating the method's stability and practical applicability for deployment under varying resource constraints.
\vspace{-0.5cm}
\begin{table}[t]
\centering
\caption{Ablation study results showing mIoU performance of DCAU-AL under different annotation budgets and learning rates}
\vspace{0.3cm}
\label{tab:dcau_ablation}
\begin{tabular}{|l|c|c|c|}
\hline
\textbf{Configuration} & \textbf{Learning Rate} & \textbf{Samples per AL Cycle} & \textbf{mIoU (\%)} \\
\hline
DCAU-Baseline      & $1\times10^{-3}$ & 50  & 0.654\\
DCAU-Baseline      & $1\times10^{-4}$ & 50  & \textbf{0.664}\\
\hline
DCAU-Extended  & $1\times10^{-3}$ & 100 & 0.632\\
DCAU-Extended  & $1\times10^{-4}$ & 100 & 0.648\\
\hline
\end{tabular}
\end{table}

\subsection*{5.4 State-of-the-art Comparison}
\vspace{-0.2cm}

To assess the effectiveness of the proposed Dynamic Class-Aware Uncertainty (DCAU-AL), we perform a comparative evaluation against state-of-the-art active learning strategies on the OpenEarthMap~\cite{xia2023openearthmap} dataset for land cover segmentation. We consider two widely used baselines: Entropy-based Uncertainty Sampling~\cite{mandalika2024segxal,yang2015multiclass}, which selects highly uncertain predictions based on pixel-wise entropy, and the Coreset~\cite{sener2017coreset} approach, which aims to select representative samples by covering the feature space. All active learning methods are evaluated under a consistent training protocol, using only 40\% of the available training data. This protocol starts with an initial labeled set and adds samples over 20 active learning cycles, with a fixed annotation budget per cycle.

As shown in Table~\ref{tab:model_miou}, the Fully Supervised Model achieves the highest mIoU of 0.670, serving as the upper bound when all training data are labeled. The Naive Active Learning Model ~\cite{mandalika2024segxal} obtains an mIoU of 0.642, while the Coreset Model ~\cite{sener2017coreset} reaches 0.648 under the same conditions. In comparison, our proposed DCAU-AL framework achieves an mIoU of \textbf{0.664}, closely approaching the fully supervised performance while utilizing only 40\% of the labeled data. Moreover, DCAU-AL maintains competitive training efficiency, requiring approximately \textbf{36 hours}, as summarized in Table~\ref{tab:model_miou}.

The qualitative predictions in Fig~\ref{fig:cycle_comparison} further demonstrate DCAU-AL's superior class-wise segmentation accuracy, particularly for buildings, roads, and water classes, where the model produces sharper boundaries and fewer misclassifications compared to baseline approaches. This enhanced class-specific performance is attributed to DCAU's dynamic sampling mechanism, which integrates class-aware uncertainty and spatial proximity through interpretable attention maps, thereby enabling more effective and targeted sample selection across the active learning process.

\vspace{-0.4cm}
\begin{table}[h!]
\centering
\caption{Segmentation Accuracy and Training Time across different models vs Proposed DCAU-AL}
\vspace{0.1cm}
\label{tab:model_miou}
\begin{tabular}{|l|c|c|}
\hline
\textbf{Model} & \textbf{mIoU Score} & {\textbf{Training Time (hrs)}} \\
\hline
Fully Supervised Model & 0.670 & 30 hr \\
Naive Active Learning Model~\cite{mandalika2024segxal} & 0.642 & 35 hr \\
Coreset Model~\cite{mittal2025realistic} & 0.648 & 38 hr \\
DCAU-AL (Ours) & {0.664} & \textbf {36 hr} \\
\hline
\end{tabular}
\end{table}

\vspace{0.1cm}

\section{Conclusions and Future Work}
In this work, we introduced the \textit{Dynamic Class-Aware Uncertainty based Active Learning} (DCAU-AL) framework, an active learning strategy that adaptively balances sample selection based on real-time class-wise performance. By re-weighting uncertainty scores with per-class IoU feedback, DCAU-AL effectively mitigates class imbalance and achieves performance comparable to fully supervised training while requiring significantly fewer annotations. A key contribution is the implementation of an adaptive thresholding mechanism that prioritizes informative yet reliable samples, enhancing stability and annotation efficiency. Extensive analysis on OpenEarth dataset validates the outperformance of the proposed DCAU-AL framework. Future work could integrate advanced Explainable AI (XAI) techniques, to improve interpretability of both segmentation outputs and sample acquisition. This would enhance transparency in human-in-the-loop annotation and provide actionable insights for domain experts.

\bibliographystyle{splncs04}

\bibliography{reference}

\end{document}